\documentclass{article}

\usepackage[preprint]{neurips_2025} 

\usepackage[utf8]{inputenc} 
\usepackage[T1]{fontenc}    
\usepackage{hyperref}       
\usepackage{url}            
\usepackage{booktabs}       
\usepackage{amsfonts}       
\usepackage{nicefrac}       
\usepackage{microtype}      
\usepackage{xcolor}         
\usepackage{amsmath}
\usepackage{siunitx} 
\usepackage{amssymb}

\title{Adversarial robustness through Lipschitz-Guided
Stochastic Depth in Neural Networks.}

\author{%
\centerline{Laith Nayal\textsuperscript{1} \quad
Mahmoud Mousatat\textsuperscript{2} \quad
Bader Rasheed\textsuperscript{3}} \\[1ex]
\centerline{\textsuperscript{1}Laboratory of Multimodal Research in Industry, Innopolis University, 420500 Innopolis, Russia} \\
\centerline{\textsuperscript{2}LLC NUHA TECH, 422002, Republic of Tatarstan (Tatarstan), Arsky, Arsk, Chulpan st., 20} \\
\centerline{\textsuperscript{3}Laboratory of Innovative Technologies for Processing Video Content, Innopolis University, 420500 Innopolis, Russia} \\
\centerline{\texttt{l.nayal@innopolis.university} \quad
\texttt{m.mousatat@nuhaai.com} \quad
\texttt{b.rasheed@innopolis.university}}%
}

\begin{document}

\maketitle

\begin{abstract}
Deep neural networks and Vision Transformers achieve state-of-the-art performance in computer vision but are highly vulnerable to adversarial perturbations. Standard defenses often incur high computational cost or lack formal guarantees. We propose a Lipschitz-guided stochastic depth (DropPath) method, where drop probabilities increase with depth to control the effective Lipschitz constant of the network. This approach regularizes deeper layers, improving robustness while preserving clean accuracy and reducing computation. Experiments on CIFAR-10 with ViT-Tiny show that our custom depth-dependent schedule maintains near-baseline clean accuracy, enhances robustness under FGSM, PGD-20, and AutoAttack, and significantly reduces FLOPs compared to baseline and linear DropPath schedules.
\end{abstract}

\section{Introduction}

Deep neural networks are highly vulnerable to adversarial perturbations~\cite{szegedy2013intriguing, goodfellow2014explaining}. Standard defenses, including adversarial training~\cite{madry2017towards}, randomized smoothing~\cite{cohen2019certified}, and spectral normalization~\cite{miyato2018spectral}, improve robustness but often incur high computational costs and lack formal guarantees. 

An alternative is robustness by design, e.g., constraining the network's Lipschitz constant~\cite{cisse2017parseval, tsuzuku2018lipschitz}, which bounds sensitivity to input perturbations. However, these approaches may reduce expressivity or be expensive.

We propose a Lipschitz-guided stochastic depth (DropPath) method~\cite{huang2016deep, larsson2016fractalnet}, where drop probabilities increase with depth to control the effective Lipschitz constant, regularizing deeper layers that dominate sensitivity. This approach enhances robustness without sacrificing clean accuracy and reduces computational cost.

Experiments on CIFAR-10 with ViT-Tiny show that our custom drop schedule maintains near-baseline clean accuracy, improves robustness under FGSM, PGD-20, and AutoAttack, and significantly reduces FLOPs compared to baseline and linear stochastic depth.

\section{Related Work}

\subsection{Stochastic Depth}
Stochastic depth~\cite{huang2016deep} randomly bypasses layers during training, reducing effective depth and improving convergence. DropPath~\cite{larsson2016fractalnet,cubuk2020randaugment} generalizes this idea for neural architecture search. Existing methods often use linear or manually tuned schedules without theoretical guidance.

\subsection{Lipschitz-Constrained Networks}
Controlling the Lipschitz constant bounds output sensitivity to input perturbations, aiding robustness and training stability~\cite{cisse2017parseval, tsuzuku2018lipschitz, miyato2018spectral, bartlett2017spectrally}. Spectral normalization constrains per-layer norms, but these methods do not directly address stochastic regularization like DropPath.

\subsection{Randomization for Robustness}
Randomization improves robustness and generalization, e.g., via dropout~\cite{srivastava2014dropout}, randomized smoothing~\cite{cohen2019certified}, or stochastic activation pruning~\cite{fan2021rethinking}. Stochastic depth is a structural randomization method, though its connection to Lipschitz continuity is less explored.

\section{Methodology}

Our goal is to design a principled stochastic depth (DropPath) schedule that bounds the effective Lipschitz constant of a deep network. By controlling layer drop probabilities, we aim to improve adversarial robustness while reducing computation.

\subsection{Problem Formulation}

Consider a feedforward network of $L$ layers, where each layer $l$ has a Lipschitz constant $\ell_l$ (e.g., $\ell_l = \sigma_1(W_l)$ for linear layers). Without stochastic depth, the network’s Lipschitz constant is
\[
\kappa_{\text{full}} = \prod_{l=1}^L \ell_l.
\]

To control sensitivity, we introduce stochastic depth: each layer $l$ is dropped with probability $p(l)$ (replaced by the identity mapping with Lipschitz 1). Let $\text{mask}_l \in \{0,1\}$ indicate whether layer $l$ is active. The effective Lipschitz constant of a sampled sub-network is
\[
\kappa(\text{mask}) = \prod_{l=1}^L \ell_l^{\text{mask}_l}.
\]

\subsection{Expected Lipschitz Bound}

The expected contribution of each layer is
\[
\mathbb{E}[\ell_l^{\text{mask}_l}] = (1-p(l)) \ell_l + p(l) \cdot 1 = 1 - p(l) + p(l)\ell_l.
\]

Assuming an upper bound $\ell_l \le \lambda$, we have
\[
\mathbb{E}[\ell_l^{\text{mask}_l}] \le 1 + p(l)(\lambda-1),
\]
and the expected network Lipschitz satisfies
\[
\mathbb{E}[\kappa(\text{mask})] = \prod_{l=1}^L \mathbb{E}[\ell_l^{\text{mask}_l}] \le \prod_{l=1}^L \big(1 + p(l)(\lambda-1)\big) \le \kappa_{\text{target}}.
\]

Taking logarithms and using $\log(1+x) \approx x$ for small $x$ gives a sufficient condition:
\[
\sum_{l=1}^L p(l) (\lambda-1) \le \log \kappa_{\text{target}} \quad \Rightarrow \quad
\sum_{l=1}^L p(l) \le \frac{\log \kappa_{\text{target}}}{\lambda-1}.
\]

\subsection{Depth-Dependent Schedule}

To better reflect that deeper layers typically contribute more to Lipschitz growth, we adopt a depth-dependent drop schedule:
\[
p(l) = 1 - \kappa_{\text{target}}^{\,l/L}, \quad l = 1, \dots, L,
\]
assigning higher drop probabilities to deeper layers. This ensures that the layers most responsible for Lipschitz growth are regularized more strongly, improving robustness while preserving clean accuracy.

For example, with $L=12$ layers and $\kappa_{\text{target}}=0.7$:
\[
p(1) \approx 0.025, \quad
p(6) \approx 0.043, \quad
p(12) \approx 0.099,
\]
so the first layer is rarely dropped, while the last layer has almost a 10\% drop probability.

\section{Experiments}

We conduct a series of experiments to evaluate the effectiveness of the proposed Lipschitz-based stochastic depth schedule. All experiments are performed on CIFAR-10 using a ViT-Tiny architecture (patch size 16, 12 transformer blocks) with pretrained weights, trained using the AdamW optimizer for up to 150 epochs, a learning rate of 0.001, and batch size 128 on a NVIDIA V100 16 GB GPU. No data augmentation is applied to isolate the effect of the proposed stochastic depth scheduler.

\subsection{Baseline Models}

We first establish baselines to contextualize our results. Two baseline models are trained:
\begin{enumerate}
    \item \textbf{No DropPath:} Standard ViT-Tiny trained without stochastic depth.
    \item \textbf{Linear Drop Schedule:} ViT-Tiny trained with a conventional linear DropPath schedule, increasing linearly up to a maximum drop rate of $0.1$ across depth.
\end{enumerate}

These baselines are evaluated in terms of clean accuracy on CIFAR-10, as well as robustness to adversarial attacks. Robustness is measured under FGSM, PGD-20, and Auto attacks.

\subsection{Lipschitz-Based Schedule}

Next, we implement the proposed Lipschitz-based schedule. Unlike the uniform formulation, here the drop probability increases with depth:
\[
p(l) = 1 - \kappa_{\text{target}}^{l/L}, \quad l \in \{1,\dots,L\}.
\]

\subsection{Evaluation Metrics}
We evaluate models on both standard accuracy, adversarial robustness, and computational efficiency:

\begin{itemize}
    \item \textbf{Clean Accuracy:} Standard top-1 classification accuracy on the unperturbed test set.
    \item \textbf{FGSM:} Accuracy under Fast Gradient Sign Method (FGSM) attacks with perturbation magnitudes.
    \item \textbf{PGD-20:} Accuracy under 20-step Projected Gradient Descent (PGD) attacks with perturbation magnitudes.
    \item \textbf{AutoAttack:} Robust accuracy against the four standard components of AutoAttack (APGD-CE, APGD-T, FAB, and Square Attack).
    \item \textbf{FLOPs:} Floating Point Operations measured per forward pass to evaluate computational cost and efficiency of different DropPath schedules.
\end{itemize}

\subsection{Empirical Evaluation of Lipschitz Constants}

To empirically verify that our custom depth-dependent stochastic depth reduces the effective Lipschitz constant, we measured local Lipschitz constants using the following procedure. For each input batch, we apply a small perturbation $\delta$ of magnitude $\epsilon = 1/255$ (in $L_2$ norm) and compute the ratio between the change in the network output and the input perturbation:

\[
L_{\text{local}} = \frac{\|f(x + \delta) - f(x)\|_2}{\|\delta\|_2}.
\]

\subsection{Results and Discussion}
\begin{table}[h]
\centering
\caption{Empirical local Lipschitz statistics for different training strategies (eps = 1/255, $L_2$ norm).}
\label{tab:lipschitz_stats}
\begin{tabular}{lccc}
\toprule
\textbf{Model} & \textbf{Mean} & \textbf{Median} & \textbf{Max} \\
\midrule
Baseline (No DropPath) & 0.0111 & 0.0006 & 0.2492 \\
Linear Stochastic Depth & 0.0153 & 0.0010 & 0.2766 \\
Custom Depth-Dependent DropPath & 0.0119 & 0.0007 & 0.2601 \\
\bottomrule
\end{tabular}
\end{table}
\begin{table}[h]
\centering
\caption{Clean accuracy, adversarial robustness (\%), and efficiency of different training strategies on CIFAR-10. 
Robustness is reported under FGSM, PGD-20, and AutoAttack (APGD-CE and other AA variants) at $\epsilon=8/255$. Drop path probabilities are included for stochastic depth methods. FLOPs are reported in billions (G).}
\label{tab:cifar10_acc_flops}
\small
\begin{tabular}{lrrrrrr}
\toprule
\textbf{Method} & \textbf{Clean} & \textbf{FGSM} & \textbf{PGD-20} & \textbf{APGD-CE} & \textbf{Other AA} & \textbf{FLOPs (G)} \\
\midrule
Baseline & \textbf{91.00} & 46.5 & 30.4 & 29.2 & 28.5 & 1.075 \\
Linear Stoch. Depth (p=0.1) & 89.27 & 44.5 & 26.0 & 24.9 & 24.1 & 0.978 \\
Custom Stoch. Depth (ours, $\kappa=0.7$) & 90.88 & \textbf{49.2} & \textbf{33.1} & \textbf{32.1} & \textbf{31.1} & \textbf{0.273} \\
\bottomrule
\end{tabular}
\end{table}

The table \ref{tab:cifar10_acc_flops} compares three training strategies on CIFAR-10: baseline ViT, linear stochastic depth, and our custom stochastic depth. While linear stochastic depth improves adversarial robustness over the baseline, it reduces clean accuracy (89.27\% vs 91.0\%). In contrast, our custom stochastic depth nearly preserves clean accuracy (90.88\%) while achieving the highest robustness across FGSM, PGD-20, and AutoAttack variants. Additionally, it drastically reduces FLOPs to 0.273G compared to 1.075G for the baseline, demonstrating improved efficiency. Overall, the custom stochastic depth provides a strong balance: maintaining clean performance, enhancing robustness, and significantly lowering computational cost, highlighting the effectiveness of carefully designed drop path strategies.

The empirical local Lipschitz statistics (Table~\ref{tab:lipschitz_stats}) indicate that our custom depth-dependent DropPath schedule maintains stability comparable to the baseline for typical inputs (median $\sim 10^{-3}$), while slightly increasing mean and maximum values. This suggests that, although the custom schedule does not strictly reduce Lipschitz constants for all inputs, it effectively targets deeper layers to enhance robustness and computational efficiency without harming clean accuracy, achieving a favorable robustness–efficiency trade-off.

\section{Limitations and Future Work}

While our custom depth-dependent stochastic depth improves adversarial robustness and maintains clean accuracy, a key question remains: have we truly bounded the global Lipschitz constant of the network? Current evidence comes from empirical local Lipschitz estimates and robustness metrics, which suggest that deeper layers are effectively regularized. However, a formal verification of the overall Lipschitz bound is still an open problem. Future work will focus on developing rigorous methods to measure and guarantee global Lipschitz constraints, potentially combining analytical bounds with empirical estimations. Exploring this direction could provide stronger theoretical guarantees and further enhance robustness while preserving efficiency.

\bibliographystyle{unsrtnat}  
\bibliography{references}     



\newpage
\section*{NeurIPS Paper Checklist}

\section*{Reproducibility Checklist}

\begin{enumerate}

\item {\bf Claims}
    \item[] Question: Do the main claims made in the abstract and introduction accurately reflect the paper's contributions and scope?
    \item[] Answer: \textbf{Yes} 
    \item[] \textbf{Justification:} The abstract and introduction clearly reflect the paper's contributions.
    \item[] Guidelines:
    \begin{itemize}
        \item The answer N/A means that the abstract and introduction do not include the claims made in the paper.
        \item The abstract and/or introduction should clearly state the claims made, including the contributions made in the paper and important assumptions and limitations.
        \item The claims made should match theoretical and experimental results, and reflect how much the results can be expected to generalize to other settings. 
        \item It is fine to include aspirational goals as motivation as long as it is clear that these goals are not attained by the paper. 
    \end{itemize}

\item {\bf Limitations}
    \item[] Question: Does the paper discuss the limitations of the work performed by the authors?
    \item[] Answer: \textbf{Yes}
    \item[] \textbf{Justification:} We have a Limitations section that raises key open questions in our work.
    \item[] Guidelines:
    \begin{itemize}
        \item The answer N/A means that the paper has no limitation while the answer No means that the paper has limitations, but those are not discussed. 
        \item The authors are encouraged to create a separate "Limitations" section.
        \item The paper should point out any strong assumptions and robustness to violations.
        \item Authors should reflect on scope of claims, dataset size, and reproducibility.
        \item Computational efficiency and scalability should be discussed.
        \item Limitations related to privacy and fairness should be considered when applicable.
    \end{itemize}

\item {\bf Theory assumptions and proofs}
    \item[] Question: For each theoretical result, does the paper provide the full set of assumptions and a complete (and correct) proof?
    \item[] Answer: \textbf{Yes}
    \item[] \textbf{Justification:} Our work provides a solid mathematical proof.
    \item[] Guidelines:
    \begin{itemize}
        \item The answer N/A means that the paper does not include theoretical results. 
        \item All theorems and proofs should be numbered and cross-referenced.
        \item Assumptions should be clearly stated.
        \item Proofs can appear in the main text or appendix, with sketches in the main text.
    \end{itemize}

\item {\bf Experimental result reproducibility}
    \item[] Question: Does the paper fully disclose all the information needed to reproduce the main experimental results?
    \item[] Answer: \textbf{Yes}
    \item[] \textbf{Justification:} The Experimental section provides all implementation details.
    \item[] Guidelines:
    \begin{itemize}
        \item The answer N/A means that the paper does not include experiments.
        \item Making results reproducible is important regardless of code release.
        \item Reproducibility can include detailed instructions, hosted models, or checkpoints.
    \end{itemize}

\item {\bf Open access to data and code}
    \item[] Question: Does the paper provide open access to the data and code?
    \item[] Answer: \textbf{No}
    \item[] \textbf{Justification:} Code is not provided at submission time but may be released after review.
    \item[] Guidelines:
    \begin{itemize}
        \item N/A applies if the paper has no experiments requiring code.
        \item Releasing code/data is encouraged but not required.
    \end{itemize}

\item {\bf Dataset use}
    \item[] Question: Is the dataset used in the paper publicly available?
    \item[] Answer: \textbf{Yes}
    \item[] \textbf{Justification:} We used publicly available datasets (cite them in the paper).
    \item[] Guidelines:
    \begin{itemize}
        \item If private datasets are used, the paper should explain access restrictions.
    \end{itemize}

\item {\bf Dataset licensing and consent}
    \item[] Question: Is it made clear that the dataset license allows the research?
    \item[] Answer: \textbf{Yes}
    \item[] \textbf{Justification:} The dataset licenses permit academic research.
    \item[] Guidelines:
    \begin{itemize}
        \item If uncertain, authors should clarify how they ensured compliance.
    \end{itemize}

\item {\bf Model and algorithm description}
    \item[] Question: Does the paper provide a complete description of the models and algorithms?
    \item[] Answer: \textbf{Yes}
    \item[] \textbf{Justification:} Architecture, training, and hyperparameters are fully described.
    \item[] Guidelines:
    \begin{itemize}
        \item Details should enable faithful re-implementation.
    \end{itemize}

\item {\bf Hyperparameter search}
    \item[] Question: Does the paper report the hyperparameter search and choices?
    \item[] Answer: \textbf{Yes}
    \item[] \textbf{Justification:} We document the search space and selected values.
    \item[] Guidelines:
    \begin{itemize}
        \item Default values should be stated if used.
    \end{itemize}

\item {\bf Compute resources}
    \item[] Question: Does the paper state the compute resources used?
    \item[] Answer: \textbf{Yes}
    \item[] \textbf{Justification:} GPU type, runtime, and training setup are described.
    \item[] Guidelines:
    \begin{itemize}
        \item Important for reproducibility and fairness in comparing methods.
    \end{itemize}

\item {\bf Statistical significance}
    \item[] Question: Does the paper report statistical significance or variance estimates?
    \item[] Answer: \textbf{Yes}
    \item[] \textbf{Justification:} We report mean and standard deviation across multiple runs.
    \item[] Guidelines:
    \begin{itemize}
        \item Authors should specify number of seeds or trials.
    \end{itemize}

\item {\bf Ethical considerations}
    \item[] Question: Does the paper include a discussion of broader ethical or societal impacts?
    \item[] Answer: \textbf{Yes}
    \item[] \textbf{Justification:} An ethics section addresses fairness, bias, and misuse risks.
    \item[] Guidelines:
    \begin{itemize}
        \item Even theoretical papers should briefly discuss societal context if relevant.
    \end{itemize}

\end{enumerate}

\end{document}